\documentclass[11pt]{article}

\usepackage{acl}
\usepackage{amssymb}
\usepackage{fvextra} 
\usepackage{tikz}
\usepackage{varwidth}   
\usepackage{placeins}
\usepackage{fontawesome5} 

\usetikzlibrary{arrows.meta,positioning}
\usepackage{times}
\usepackage{multirow}
\usepackage{latexsym}
\usepackage[T1]{fontenc}
\usepackage[utf8]{inputenc}
\usepackage{microtype}
\usepackage{graphicx}
\usepackage{tabularx}
\usepackage{fancyvrb}

\usepackage{array}

\usepackage{inconsolata}
\usepackage{booktabs} 
\usepackage{amsmath}
\usepackage{float} 
\usepackage{algorithm} 
\usepackage{algpseudocode} 
\algrenewcommand\algorithmicrequire{\textbf{Inputs:}}
\algrenewcommand\algorithmicensure{\textbf{Outputs:}}
\newcommand{\Clip}{\operatorname{clip}}

\newcommand{\Ent}{\mathcal{H}}

\title{GRASP LoRA: GRPO Guided Adapter Sparsity Policy for Cross Lingual Transfer}

\author{
  Besher Hassan$^{1}$, Xiuying Chen$^{1}$\thanks{Corresponding author} \\
  $^{1}$ Mohamed bin Zayed University of Artificial Intelligence \\
  \texttt{\{besher.hassan, xiuying.chen\}@mbzuai.ac.ae}
}

\begin{document}
\maketitle



\begin{abstract}
Parameter efficient fine tuning is a way to adapt LLMs to new languages when compute or data are limited, yet adapter pipelines usually choose a global prune ratio by grid search. This practice is computationally expensive and development set intensive, since it repeats training,
freezes sparsity, and misses fractional optima. We introduce GRASP LoRA (GRPO Guided Adapter Sparsity Policy), which treats global sparsity as a learnable control variable. A GRPO controller interleaves with training, periodically probing candidate prune ratios on a small micro development set and updating a single global prune ratio online from its reward signal. It operates on merged source and target LoRA adapters on a frozen backbone and replaces grid search with one controller run that learns a prune ratio, followed by a single final merge and prune fine tuning run with pruning fixed to that ratio. On cross lingual transfer from English into Arabic and Chinese, including XL-Sum summarization and MLQA extractive question answering with Llama 3 8B, GRASP LoRA improves semantic faithfulness, content coverage, and answer quality over strong target only and merge and prune baselines. It reduces end to end runtime by multiple times relative to grid search, lowers reliance on large development sets, and makes adapter reuse practical for low resource deployment. \href{https://github.com/besherhasan/GRASP-LoRA/tree/main}{\faGithub\ GRASP LoRA}.

\end{abstract}

\section{Introduction}

Parameter efficient fine tuning significantly reduce the computational load, memory, and data required for adaptation by attaching small, trainable adapter modules to a frozen model, instead of updating all weights \citep{li2021prefixtuning,lester2021prompttuning,benzaken2022bitfit,liu2022ia3}. LoRA perfectly illustrates this approach: low rank updates are inserted at key projection sites, allowing one model to support a library of task or language adapters \citep{hu2022lora}. Adapter composition further improves reusability by combining resource rich language adapters with low resource target adapters in resource constrained environments \citep{pfeiffer2020adapterhub,pfeiffer2021adapterfusion,pfeiffer2020madx,huang2024lorahub}. However, simple composition alone is not enough. When merging multiple modules, the parameters that are important to the source language can affect the target language, and the quality of the transfer depends on the capacity retained versus removed during the adaptation process. Therefore, choosing the level of overall sparsity is a crucial decision that impacts performance and efficiency, as well as reproducibility and distribution.

In contemporary merge and prune strategies, workflows generally employ a fixed global prune ratio, determined by a coarse grid search at a few pre selected points \citep{adaptivelora2025}. Each candidate requires a full training run and using the full development set, which is an unnecessary drain on time and energy. Even after a value is chosen, it stays fixed, while the model and many other components continues to drift throughout training. This choice is not particularly robust. A plethora of grid points go unchecked, and repeated pruning shocks are likely to interact poorly with optimizer momentum. Moreover, selection for dev cost heavily biases the outcome to the peculiarities of a small dev set. learning and discovering the sparsity from the task feed back still not explored yet. The benefits of such an approach are particularly salient in resource limited, cross lingual settings.

We argue that the values related to sparsity should be learned rather than empirically searched. Rather than treating the prune ratio as a fixed hyperparameter, we reframe it as a control variable that is learnable and optimized in relation to task feedback during training. GRASP LoRA incorporates a lightweight controller that operates in parallel with the optimizer: the controller adjusts and learns the global prune ratio based from a micro development signal \citep{grpo,cppo,scafgrpo}. Therefore the method is compatible with PEFT codebases: it state that no changes to the backbone architecture, no new loss, and only a few lightweight scoring and masking calls are necessary \citep{hu2022lora,li2021prefixtuning,lester2021prompttuning,benzaken2022bitfit,liu2022ia3}. In our experiments, learning sparsity improves transfer quality of the model, while providing the same level of efficiency.

Our contributions consist of three parts. First, we propose GRASP LoRA, which replaces grid search with a learnable global pruning rate p, which is optimized in real time during training. 
Second, we demonstrate that this learnable sparsity pipeline significantly improves the selection efficiency of \(p\). On the XL-Sum and MLQA datasets, this pipeline finds the appropriate pruning level with only one controller iteration and one final merge pruning step, reducing runtime by 3.90x to 7.45x compared to the standard 8 point grid search. 
Third, we show that the resulting fractional pruning rate can build better models. By implementing cross lingual transfer from English to Arabic and Chinese, we improve the summarization quality of XL-Sum and the question answering extraction performance of MLQA. This outperforms baseline models using only target data and merge, as well as the optimal pruning ratio using grid search.

\section{Related Work}

\paragraph{PEFT and adapter composition.}
 LoRA is one such method, inserting low rank update matrices into selected projection layers of the base model \citep{hu2022lora}. Other PEFT variants include prefix tuning, prompt tuning, BitFit, and IA$^{3}$, which trade off flexibility and parameter count in different ways \citep{li2021prefixtuning,lester2021prompttuning,benzaken2022bitfit,liu2022ia3}. Adapter composition frameworks such as AdapterHub, AdapterFusion, and MAD-X support reusing and combining trained adapters for new tasks or languages without retraining them from scratch \citep{pfeiffer2020adapterhub,pfeiffer2021adapterfusion,pfeiffer2020madx} Parameter space model fusion methods like TIES Merging combine separately fine tuned models without joint training, which can reduce noise during composition \citep{yadav2023tiesmerging}. These works focus on how to select and combine adapters, but sparsity is usually set by fixed heuristics rather than being learned from training signals.

\paragraph{Pruning for LLMs.}

Pruning helps LLMs save memory and compute by removing parameters that contribute little to model predictions. SparseGPT, Pruning by Weights and Activations, and LLM-Pruner apply magnitude based methods and other forms of structure pruning to identify such least contribution weights for dense LLMs \citep{frantar2023sparsegpt,wanda,ma2023llmpruner}. In PEFT, adapter pruning helps to remove LoRA entries post merging that are deemed less significant based on a global deletion rate that is selected via grid search \citep{adaptivelora2025}. While LoRAPrune and LoRADrop \citep{loraprune2024,loradrop2024} employ some forms of structural or outputs sensitivity to prune some LoRA modules, they still adhere to fixed quantities of the sparsity schedules. Of the compression methods, AMC learns pruning policies and HAQ learns quantization policies, framing model compression as a sequential decision making problem \citep{amc,wang2019haq}. In a similar vein, recent work frames LLM latency and sparsity trade offs in terms of policy search \citep{rap2025}.  In contrast, rather than learning a separate action per layer or fixing the global ratio by grid search, we learn a single global prune ratio $p$ online from micro dev rewards using a GRPO style controller with mean anchoring and an entropy bonus \citep{grpo,cppo,scafgrpo}.

\paragraph{Cross lingual transfer.}

Multilingual pre training and language specific parameter separation facilitate cross lingual transfer. Encoder decoder models such as mBART and mT5 can be used to transfer the task skills across languages, especially when jointly configured and using language adapters \citep{tang2020mbart,xue2021mt5}. Adapter based approaches enhance modularity. AdapterHub supports reuse, while AdapterFusion and MAD-X can fuse adapters and route them to new targets \citep{pfeiffer2020adapterhub,pfeiffer2021adapterfusion,pfeiffer2020madx}. For LoRA, LoRAHub explores compositions of task adapters without retraining \citep{huang2024lorahub}. Furthermore, when combining LoRA adapters for source and target languages, it is important to consider how much memory to conserve during training. Previous methods avoid pruning or change the prune ratio of grid search \citep{adaptivelora2025}. In contrast, we learn the ratio through online micro dev set during policy search, performing one merge and one pruning. This combination of sparsity and utility reduces runtime while maintaining or improving quality of the generation using small dev set. Furthermore, the micro development protocol limits the overload of the development dataset and generates verifiable policy records.

\section{Method}
\label{sec:method}

\begin{figure*}[t]
  \centering
  \includegraphics[width=\textwidth]{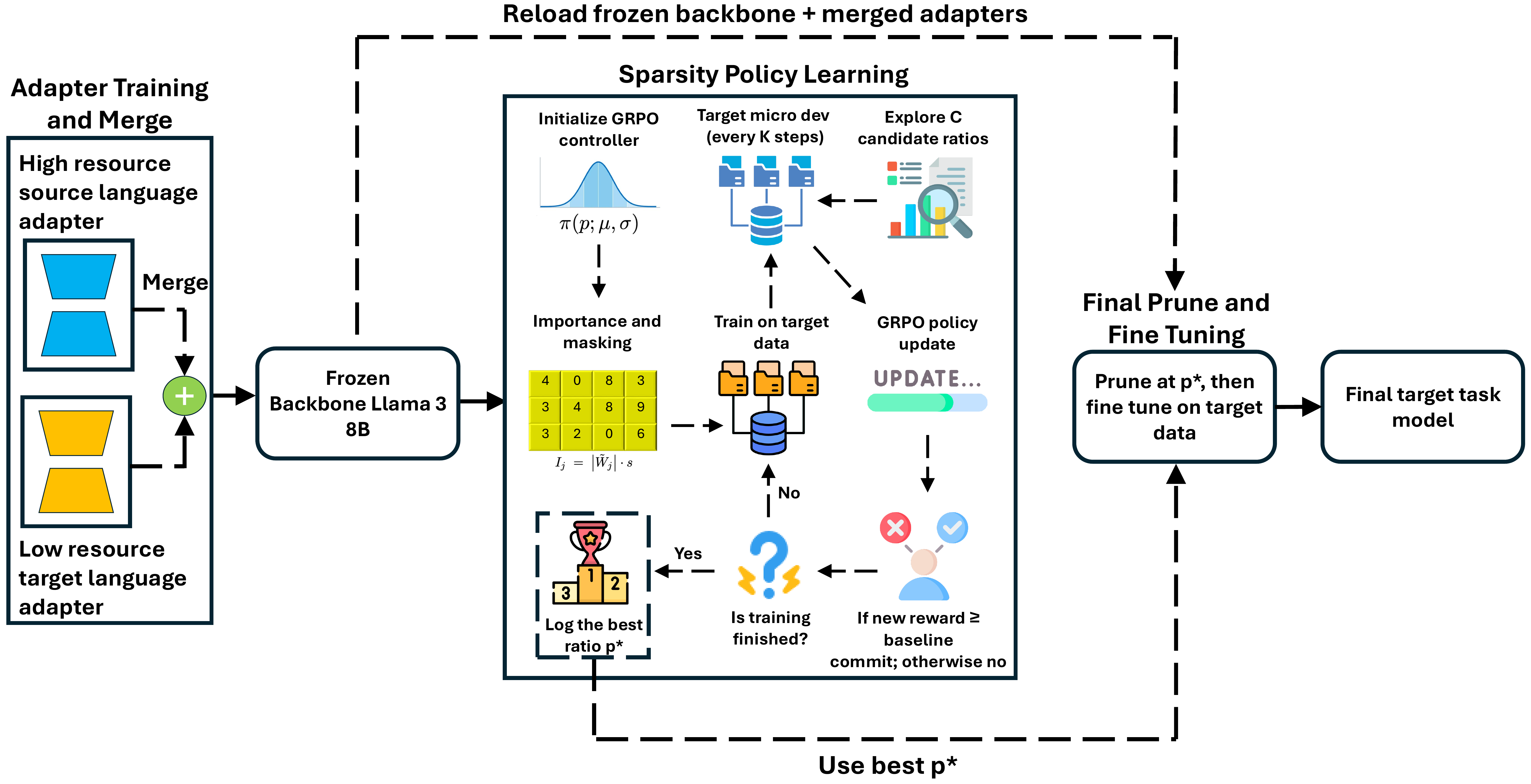}
\caption{GRASP LoRA pipeline in three phases: merge source and target LoRA adapters on a frozen backbone, learn prune ratio $p^\star$ from micro dev feedback with a GRPO controller, then prune once at $p^\star$ and fine tune on target data to produce the final target task model.}

  \label{fig:grpo-controller}
\end{figure*}

\subsection{Problem Formulation}
\label{subsec:problem}

We study cross lingual transfer using a pretrained backbone $F(x;\theta)$, where $x$ is an input token sequence and $\theta$ are the frozen backbone parameters, and we adapt it with LoRA adapters. 

Let each adapter $a$ at a projection site provide LoRA factors $(A_a, B_a)$, with $\Delta W_a = B_a A_a$.
We merge adapters by summing their low rank updates:
\begin{equation}
\Delta W_{\mathrm{merge}} = \sum_a \Delta W_a = \sum_a (B_a A_a).
\label{eq:merge}
\end{equation}

Let $h$ denote the input hidden states to this projection layer, and let $W$ be the corresponding frozen base weight matrix. The projection output becomes:
\[
h(W + \Delta W_{\mathrm{merge}})^\top \;=\; hW^\top + h\Delta W_{\mathrm{merge}}^\top,
\]
where $(\cdot)^\top$ denotes transpose. Thus, the LoRA contribution is the additive term $h\Delta W_{\mathrm{merge}}^\top$.

We then apply binary masks to the merged LoRA parameters. Let $\{\tilde{W}^{(t)}\}_{t=1}^T$ denote the collection of merged LoRA parameter tensors across layers and projection sites, where $T$ is the total number of such tensors and tensor $t$ has $d_t$ entries. For a prune ratio $p\in[0,1]$ reported as $100p$ in percent, we construct a mask $M^{(t)}(p)\in\{0,1\}^{d_t}$ for each tensor and define the effective sparse adapters:
\[
\hat{W}^{(t)}(p)=M^{(t)}(p)\odot \tilde{W}^{(t)} .
\]

A prune ratio \(p\) means we target pruning approximately a fraction \(p\) of entries within each tensor.
For tensor \(t\) with \(d_t\) entries, let \(k_t=\lfloor p\,d_t\rfloor\) be the number of entries to prune, and let $\tau_t(p)$ be the pruning threshold, defined as the $k_t$-th smallest value of the importance scores $I^{(t)}$ defined in Sec.~\ref{subsec:importance}.
For each tensor $t$ and entry index $j \in \{1,\dots,d_t\}$, we prune by thresholding:
\[
\begin{aligned}
P^{(t)}_j(p) &= \mathbb{1}\!\left[I^{(t)}_j \le \tau_t(p)\right],\\
M^{(t)}_j(p) &= 1 - P^{(t)}_j(p).
\end{aligned}
\]
where $\mathbb{1}[\cdot]$ is the indicator function. Here $P^{(t)}_j(p)=1$ indicates that entry $j$ is pruned and $P^{(t)}_j(p)=0$ indicates it is kept; the mask $M^{(t)}_j(p)$ is the corresponding keep mask  where $M^{(t)}_j(p)=1$ keeps and $M^{(t)}_j(p)=0$ prunes.

Our goal is to choose a global prune ratio $p$ and the induced mask $M(p)$ that yields strong target language validation performance.

\subsection{Adapter Training and Merge}
\label{subsec:adapter-merge}

We first train two task-specific LoRA adapters on a shared frozen backbone: a high resource English adapter on the English training set, and a low resource target adapter on the target language training set. Both adapters use the same LoRA configuration and are injected into the same projection sites. After training, we save the two adapter checkpoints. For sparsity policy learning and final prune and fine tuning, we reload the frozen backbone and reconstruct a merged initialization by summing the two adapters’ low rank updates at each site Eq.~\eqref{eq:merge}. Importantly, final prune and fine tuning starts from this pre controller merged initialization.

\subsection{Sparsity Policy Learning}
\label{subsec:controller}

The sparsity policy learning phase corresponds to the middle panel in Figure~\ref{fig:grpo-controller}. It starts from the pre-controller merged adapters constructed in the adapter training and merge phase and continues target language fine tuning with controller rounds interleaved every \(K\) steps. The nine boxes in the middle panel map to the following components.

\paragraph{Reward definition.}
We use the negative micro dev loss as the controller reward:

\begin{equation*}
\tilde{R}(p) \;=\; -\,\ell_{\text{val}}(p).
\end{equation*}

where $\ell_{\text{val}}(p)$ is evaluated on a fixed micro dev subset under the masked adapters $\hat{W}(p)$.
We also use a relative reward with respect to the committed ratio $p_{\text{curr}}$:

\begin{equation*}
R(p) \;=\; \tilde{R}(p) - \tilde{R}(p_{\text{curr}}).
\end{equation*}

so the commit rule $R(p_{\text{best}})\ge 0$ means the candidate does not worsen micro dev loss.

\paragraph{Initialize GRPO controller.}
Before training, we fix the controller hyperparameters \(p_{\min}\), \(p_{\max}\), \(p_{\text{init}}\), \(K\), \(C\), \(m\), \(\eta\), \(\beta\), \(\tau\), and \(\Delta_{\max}\), and initialize the Gaussian policy \(\pi_{\mu,\sigma}\) over the prune ratio \(p\) with \(\mu=p_{\text{init}}\) and \(\sigma=(p_{\max}-p_{\min})/6\). We also set the initial committed ratio to \(p_{\text{curr}}=p_{\text{init}}\).

\paragraph{Importance and masking.}
\label{subsec:importance}
We rank LoRA parameters by a scaled magnitude score. Following prior importance scoring for pruning \citep{wanda,adaptivelora2025}, for each scalar adapter entry $j$ we define
\begin{equation*}
I_j \;=\; \big|\tilde{W}_j\big| \cdot s .
\end{equation*}

where \(\tilde{W}_j\) is the \(j\)-th scalar entry of the merged LoRA parameter tensors \(\tilde{W}\). These are the merged LoRA weights at each injection site. Pruning is applied independently per LoRA tensor, typically the \(A\) and \(B\) matrices.
The scale \(s\) is a single global scalar estimated once from the micro dev inputs as the mean \(\ell_2\) norm of the token embedding vectors.
Because \(s\) is constant across all entries, multiplying by \(s\) does not change the ordering of importance scores, so the induced per tensor top-$k$ masks are identical to using \(|\tilde{W}_j|\) alone.

Given a candidate prune ratio $p$, we construct masks independently for each LoRA tensor.
We denote this procedure as build mask, implemented by per tensor order statistic thresholding.
For a tensor $t$ with $d_t$ entries, we set $k_t=\lfloor p\,d_t\rfloor$ and obtain a threshold $\tau_t(p)$ as the $k_t$-th smallest importance value, the $k_t$-th order statistic of the flattened scores.
We prune entries with $I^{(t)} \le \tau_t(p)$, so ties at the threshold may prune slightly more than $k_t$ entries.

\paragraph{Train on target data.}
Between GRPO rounds, we continue fine tuning the merged adapters on target language data using the currently committed mask \(M(p_{\text{curr}})\). After each optimizer step we reapply the mask so pruned coordinates stay zero, and newly pruned entries have their optimizer state cleared at commit time. This implements the ``Train on target data'' box in Figure~\ref{fig:grpo-controller}.

\paragraph{Target micro dev (every \(K\) steps).}
Controller rounds occur every \(K\) optimizer steps. In each round, we evaluate a fixed micro dev slice of \(m\) target-language examples that supplies the reward signal; the same slice is reused for the baseline and for all \(C\) candidate prune ratios in that round. We first evaluate the currently committed masked model on this slice to obtain the baseline reward
\[
\tilde{R}_{\text{base}} \;=\; \tilde{R}(p_{\text{curr}}).
\]
The first round starts from \(p_{\text{curr}}=p_{\text{init}}\), so \(\tilde{R}_{\text{base}}=\tilde{R}(p_{\text{init}})\). We then evaluate each candidate \(p_i\) on the same micro dev slice and store its relative reward
\[
R(p_i) \;=\; \tilde{R}(p_i) - \tilde{R}_{\text{base}}.
\]
For brevity, we denote $R_i \equiv R(p_i)$ for the $i$-th candidate in the current round.
After scoring, we commit only if the best candidate \(p_{\text{best}}=\arg\max_{i} R(p_i)\) satisfies \(R(p_{\text{best}})\ge 0\); on commit, \(p_{\text{curr}}\) moves toward \(p_{\text{best}}\) with a step clipped by \(\pm\Delta_{\max}\) and clamped to \([p_{\min},p_{\max}]\). Otherwise \(p_{\text{curr}}\) and its mask are kept unchanged for the next round.

\paragraph{Explore \(C\) candidate ratios.}
For each candidate, we sample an unconstrained value \(z_i \sim \mathcal{N}(\mu,\sigma^2)\) and set the prune ratio \(p_i=\mathrm{clamp}(z_i,p_{\min},p_{\max})\), so values outside \([p_{\min},p_{\max}]\) are mapped to the nearest bound. We enforce a small minimum \(\sigma\) for numerical stability. In each round we evaluate the \(C\) candidates on the same fixed micro dev slice of \(m\) target examples. During probing, we disable gradient computation, temporarily apply the candidate mask for evaluation, and then restore the original parameters. The \(\Delta_{\max}\) bound is applied only when committing a winning candidate (no commit if all rewards are negative), not during probing.

\paragraph{GRPO policy update.}
We update the Gaussian policy $\pi_{\mu,\sigma}$ by maximizing the objective:
\begin{equation}
\label{eq:policy_objective}
\begin{aligned}
J(\mu,\sigma) \;=\;& \mathbb{E}_{p\sim \pi_{\mu,\sigma}}\!\left[R(p)\right]
- \frac{\beta}{2}\,(\mu - p_{\text{curr}})^2 \\
& + \tau\,\Ent\!\big(\pi_{\mu,\sigma}\big).
\end{aligned}
\end{equation}
where $\Ent(\pi_{\mu,\sigma})$ is the entropy of $\mathcal{N}(\mu,\sigma^2)$ and promotes exploration.

We draw $z_i \sim \mathcal{N}(\mu,\sigma^2)$ for $i=1,\dots,C$ and set $p_i=\mathrm{clamp}(z_i,p_{\min},p_{\max})$.
We denote the relative reward of candidate $p_i$ by $R_i \equiv R(p_i) = \tilde{R}(p_i) - \tilde{R}(p_{\text{curr}})$.
For variance reduction, we use centered advantages:
\[
A_i \;=\; R_i - \frac{1}{C}\sum_{j=1}^{C} R_j.
\]
For the score function (likelihood ratio) gradients, where $\theta = (\mu,\sigma)$, we use the identity:
\[
\nabla_{\theta}\,\mathbb{E}_{p\sim \pi_{\theta}}[R(p)]
=
\mathbb{E}_{p\sim \pi_{\theta}}\!\left[R(p)\,\nabla_{\theta}\log \pi_{\theta}(p)\right],
\]
and estimate this gradient with $C$ Monte Carlo samples.

We use the Gaussian score $\nabla_{\theta}\log \pi_{\mu,\sigma}(z_i)$ for $\mathcal{N}(\mu,\sigma^2)$ and treat the clamp as a truncation approximation.

We update the Gaussian policy parameters $(\mu,\sigma)$ using the corresponding score function estimators:
\[
\begin{aligned}
g_\mu \;&=\; \frac{1}{C}\sum_{i=1}^{C} A_i\,\frac{(z_i-\mu)}{\sigma^2},\\
g_\sigma \;&=\; \frac{1}{C}\sum_{i=1}^{C} A_i\,\frac{(z_i-\mu)^2-\sigma^2}{\sigma^3}.
\end{aligned}
\]

The mean update includes a proximal term toward the committed ratio $p_{\text{curr}}$ via mean anchoring.
To prevent premature collapse of exploration, we implement the entropy term as a small exploration bonus that increases $\sigma$ by a constant offset each round.
We also floor $\sigma$ at $10^{-3}$ for numerical stability:
\begin{equation}
\label{eq:surrogate}
\begin{aligned}
\mu &\gets \mu + \eta g_\mu - \eta\beta(\mu-p_{\text{curr}}),\\
\sigma &\gets \max\big(10^{-3},\, \sigma + \eta g_\sigma + \tau\big),\\
\mu &\gets \operatorname{clamp}_{[p_{\min},p_{\max}]}(\mu).
\end{aligned}
\end{equation}

\paragraph{Commit rule.}
We allow \(p\) to increase or decrease, but cap the per-update change by \(\Delta_{\max}\) to limit abrupt sparsity shifts and mask thrashing.
Let \(i^\star = \arg\max_{i} R_i\) and \(p_{\text{best}} = p_{i^\star}\).
We commit only if \(R_{i^\star} \ge 0\), meaning the best candidate does not worsen the micro dev loss relative to the current setting.
If the condition holds, we update the committed ratio by a bounded, clamped step:
\begin{align}
\Delta p \;&=\; \Clip\!\big(p_{\text{best}}-p_{\text{curr}}, -\Delta_{\max}, \Delta_{\max}\big), \label{eq:delta_p}\\
\tilde{p} \;&=\; p_{\text{curr}} + \Delta p, \label{eq:p_tilde}\\
p_{\text{curr}} \;&\leftarrow\; \operatorname{clamp}_{[p_{\min},p_{\max}]}\!\big(\tilde{p}\big). \label{eq:commit}
\end{align}
We then rebuild the pruning mask at \(p_{\text{curr}}\), zero newly pruned coordinates, and reset optimizer state at newly pruned indices.
If no candidate improves on the baseline, we keep \(p_{\text{curr}}\) unchanged.

\paragraph{Training termination.}
Sparsity policy learning runs for a fixed budget of \(E\) epochs equivalently \(T\) optimizer steps, with one controller round every \(K\) steps.

\paragraph{Logging and selecting \texorpdfstring{$p^\star$}{p*}.}
At each GRPO round, we record the current committed ratio \(p_{\text{curr}}\) and its micro dev evaluation \(\tilde{R}(p_{\text{curr}})\), together with the sampled candidate ratios \(\{p_i\}_{i=1}^{C}\) and their relative rewards \(R_i=\tilde{R}(p_i)-\tilde{R}(p_{\text{curr}})\) (plus the commit decision).
After sparsity policy learning ends, we select \(p^\star\) by scanning the log and choosing the ratio that achieves the highest implied micro dev reward across all rounds, computed from the recorded baseline \(\tilde{R}(p_{\text{curr}})\) and the relative rewards \(R_i\).

\subsection{Final Prune and Fine Tuning}
\label{subsec:final-prune}

After sparsity policy learning ends, we select the best prune ratio \(p^\star\) from the controller logs.
We then launch a separate final run that reloads the frozen backbone and the merged adapters from the adapter training and merge phase, that is, the pre-controller merged initialization.
We construct the sparsity masks once at \(p^\star\) using the importance rule in Sec.~\ref{subsec:importance}, and then fine tune on the target language training set while keeping the mask fixed throughout training.
The remaining target dev examples, disjoint from the micro dev slice, are used only for early stopping.

\section{Experiments}

\subsection{Datasets}
We evaluate cross lingual transfer on XL-Sum summarization \citep{hasan2021xlsum} and MLQA extractive QA \citep{lewis2020mlqa}, transferring from English to Arabic and Chinese. Table~\ref{tab:data-sizes} reports split sizes, where English uses a much larger training set than Arabic and Chinese to simulate a high resource source and low resource targets. For each target language, we create a fixed micro dev split of 16 examples used only for controller rewards; the standard dev split is separate and used only for early stopping.

\begin{table}[H]
\centering
\footnotesize
\setlength{\tabcolsep}{4pt}
\renewcommand{\arraystretch}{1.05}
\begin{tabular}{llrrrr}
\toprule
\textbf{Dataset} & \textbf{Lang} & \textbf{Train} & \textbf{Dev} & \textbf{Micro dev} & \textbf{Test} \\
\midrule
\multirow{3}{*}{XL-Sum} & En & 10000 & 1000 & \textsc{n/a} & 1000 \\
                        & Ar & 50    & 50   & 16           & 100  \\
                        & Zh & 50    & 50   & 16           & 100  \\
\midrule
\multirow{3}{*}{MLQA}   & En & 3030  & 315  & \textsc{n/a} & 315  \\
                        & Ar & 50    & 50   & 16           & 100  \\
                        & Zh & 50    & 50   & 16           & 100  \\
\bottomrule
\end{tabular}
\caption{Dataset split sizes.}
\label{tab:data-sizes}
\end{table}

\newcommand{\adl}[1]{{Adaptive LoRA}{\scriptsize~#1}}
\newcommand{\grasp}[1]{{GRASP LoRA}{\scriptsize~#1}}

\newcommand{\blockheadJoint}[1]{%
  \addlinespace[0.4em]
  \multicolumn{9}{l}{\textbf{#1} {\scriptsize}}\\[-0.35em]
  \cmidrule(lr){1-9}
}

\begin{table*}[t]
\centering
\normalsize
\setlength{\tabcolsep}{4.5pt}
\renewcommand{\arraystretch}{1.05}
\resizebox{\textwidth}{!}{%
\begin{tabular}{lcccccccc}
\hline
 & \multicolumn{4}{c}{\textbf{XL-Sum}} & \multicolumn{4}{c}{\textbf{MLQA}} \\
\cmidrule(lr){2-5}\cmidrule(lr){6-9}
\textbf{Type}
& \textbf{Pruning} & \textbf{BERTScore-F1} & \textbf{BLEU-4} & \textbf{ROUGE-L}
& \textbf{Pruning} & \textbf{BERTScore-F1} & \textbf{EM} & \textbf{Token F1} \\
\hline

\blockheadJoint{Without Pruning Arabic}

Zero shot (Ar)
& \textemdash{} & 69.08 & 2.05 & 10.60
& \textemdash{} & 65.71 & 0.00 & 11.00 \\

LoRA (Ar)
& \textemdash{} & 74.33 & 4.46 & 18.17
& \textemdash{} & 84.07 & 29.00 & 43.84 \\

TIES merge (Ar+En)
& \textemdash{} & 69.59 & 2.19 & 11.99
& \textemdash{} & 67.19 & 2.00 & 13.28 \\

LoRAHub (Ar+En)
& \textemdash{} & 73.76 & 5.10 & 16.85
& \textemdash{} & 83.46 & 29.00 & 42.79 \\

\adl{Merge} (Ar+En)
& \textemdash{} & 74.32 & 4.46 & 18.15
& \textemdash{} & 84.80 & 36.00 & 47.77 \\

\blockheadJoint{Best grid search over [10\% -- 80\%] Arabic}

\adl{Merge + pruning} (Ar+En)
& 70
& $74.96 \pm 0.25$
& $6.18 \pm 0.29$
& $20.19 \pm 0.52$
& 40
& $85.62 \pm 0.09$
& $38.33 \pm 0.58$
& $50.29 \pm 0.39$ \\

\blockheadJoint{Learned prune ratio Arabic}

\textbf{Ours \grasp{Merge + pruning} (Ar+En)}
& \textbf{67.49}
& $\textbf{75.84} \pm 0.13$
& $\textbf{7.93} \pm 0.10$
& $\textbf{22.32} \pm 0.31$
& \textbf{48.97}
& $\textbf{86.18} \pm 0.27$
& $\textbf{41.00} \pm 2.08$
& $\textbf{52.51} \pm 0.98$ \\

\blockheadJoint{Without Pruning Chinese}

Zero shot (Ch)
& \textemdash{} & 26.48 & 9.39 & 20.87
& \textemdash{} & 7.12 & 0.00 & 6.95 \\

LoRA (Ch)
& \textemdash{} & 31.09 & 13.73 & 28.15
& \textemdash{} & 64.15 & 42.00 & 47.17 \\

TIES merge (Ch+En)
& \textemdash{} & 26.23 & 9.37 & 21.30
& \textemdash{} & 11.28 & 0.00 & 13.51 \\

LoRAHub (Ch+En)
& \textemdash{} & 26.42 & 10.35 & 24.16
& \textemdash{} & 65.11 & 44.00 & 48.10 \\

\adl{Merge} (Ch+En)
& \textemdash{} & 32.01 & 14.33 & 29.14
& \textemdash{} & 67.45 & 46.00 & 49.63 \\

\blockheadJoint{Best grid search over [10\% -- 80\%] Chinese}

\adl{Merge + pruning} (Ch+En)
& 50
& $32.00 \pm 0.36$
& $14.81 \pm 0.26$
& $29.50 \pm 0.41$
& 10
& $69.30 \pm 1.37$
& $46.50 \pm 0.71$
& $50.93 \pm 1.88$ \\

\blockheadJoint{Learned prune ratio Chinese}

\textbf{Ours \grasp{Merge + pruning} (Ch+En)}
& \textbf{56.94}
& $\textbf{33.62} \pm 0.16$
& $\textbf{16.54} \pm 0.15$
& $\textbf{30.95} \pm 0.26$
& \textbf{23.73}
& $\textbf{71.28} \pm 0.29$
& $\textbf{48.00} \pm 0.58$
& $\textbf{51.60} \pm 0.29$ \\

\hline
\end{tabular}}
\caption{Joint results on XL-Sum and MLQA for Arabic and Chinese. Each dataset block reports pruning ratio and core metrics. All scores are on a 0 to 100 scale. Entries of the form $a \pm b$ report mean $a$ and standard deviation $b$ over three seeds, and we report mean and standard deviation only for the best baseline and the best overall method in the table.
}
\label{tab:joint-xlsum-mlqa}
\end{table*}

\subsection{Baselines}
We compared the following settings for each target language Arabic and Chinese and for each dataset XL-Sum and MLQA: (1) Zero shot, using the base Llama 3 8B model without target language fine tuning. (2) Target only LoRA, the standard LoRA fine tuning on the small target-language training set. (3) Adapter merging without pruning, merging the English and target language adapters without pruning. We considered three merging strategies: (i) directly merging the LoRA adapters for English and the target language referred to as adaptive LoRA merging, followed by finetuning on target language data; (ii) using TIES adapter merging with the English and target language adapters \citep{yadav2023tiesmerging}; and (iii) merging the English and target language adapters using LoRAHub \citep{huang2024lorahub}. 4) Merging and pruning using grid search. Specifically, the adaptive LoRA Parameter Pruning Merge Baseline \citep{adaptivelora2025} method is used, which performs a grid search over prune ratios \(p \in \{0.10, 0.20, \dots, 0.80\}\). For each \(p\)-value, a full training run is required and the best results are selected based on the development set.

\subsection{Implementation}
\label{subsec:impl}
All experiments use a frozen Llama 3 8B backbone on one NVIDIA A100 GPU. LoRA adapters are injected into the Q and V projection matrices for all runs; rank and LoRA hyperparameters are fixed across methods, and all runs share the same training hyperparameters in Appendix~\ref{app:hyperparams}. For GRASP LoRA, all controller hyperparameters are fixed across runs, and we select \((\beta,\tau)\) per task and language from \(\{(0.04,0.02),(0.04,0.01),(0.05,0.01)\}\): Arabic XL-Sum uses \((0.04,0.02)\), Chinese XL-Sum uses \((0.04,0.01)\), Arabic MLQA uses \((0.04,0.01)\), and Chinese MLQA uses \((0.05,0.01)\). These settings achieve the highest GRASP LoRA scores reported in Table~\ref{tab:joint-xlsum-mlqa}. We use seeds \(\{42,1337,9001\}\) for GRASP LoRA and the strongest baselines in Table~\ref{tab:joint-xlsum-mlqa} to compute mean and standard deviation. We use a maximum input length of 2200 tokens across tasks, with generation limits of 128 new tokens for Arabic XL-Sum, 96 for Chinese XL-Sum, and 64 for MLQA Arabic and Chinese. Prompt templates are reported in Appendix~\ref{app:prompts}.

\subsection{Metrics}
For \textbf{XL-Sum} we report BERTScore-F1, BLEU-4, and ROUGE-L.
For \textbf{MLQA} we report BERTScore-F1 on predicted answer strings, Exact Match (EM), and token level QA F1.
Full metric configuration, including ROUGE granularity and SacreBLEU tokenization, is in Appendix~\ref{app:metric-details}.
Additional metric results are reported in Appendix~\ref{app:additional-results} Table~\ref{tab:appendix-extra-metrics}.

\subsection{Main Results}
Table~\ref{tab:joint-xlsum-mlqa} reports test results on XL-Sum summarization and MLQA extractive QA for Arabic and Chinese.
Across both tasks and both target languages, GRASP LoRA achieves the best overall performance, while selecting the pruning level automatically rather than via grid search.

\paragraph{XL-Sum summarization.}
GRASP LoRA outperforms the best grid searched merge and prune baseline on both languages, improving Arabic by +0.88 BERTScore-F1, +1.75 BLEU-4, and +2.13 ROUGE-L, and improving Chinese by +1.62 BERTScore-F1, +1.73 BLEU-4, and +1.45 ROUGE-L.
It learns fractional ratios of $p^\star{=}67.49\%$ (Arabic) and $p^\star{=}56.94\%$ (Chinese), whereas grid search is restricted to discrete pruning levels.

\paragraph{MLQA extractive QA.}
GRASP LoRA also yields the strongest QA results.
Relative to the best grid baseline, it improves Arabic by +0.56 BERTScore-F1, +2.67 EM, and +2.22 token F1, and improves Chinese by +1.98 BERTScore-F1, +1.50 EM, and +0.67 token F1 (Table~\ref{tab:joint-xlsum-mlqa}).
The learned prune ratios are task dependent, with $p^\star{=}48.97\%$ for Arabic MLQA and $p^\star{=}23.73\%$ for Chinese MLQA.

\paragraph{Qualitative analysis.}
 Figure~\ref{fig:qual-panel} in Appendix~\ref{app:additional-results} shows representative examples: on XL-Sum Arabic, GRASP LoRA preserves key factual details (Tripoli as Libya’s capital) that baselines omit, and on MLQA Arabic it extracts the correct year (2013) where other methods predict incorrect spans.

\section{Analysis \& Discussion}

\subsection{Stabilizing Online Sparsity Learning}
\label{sec:stab}
We use three mechanisms to stabilize online sparsity learning: (i) mean anchoring, (ii) an entropy bonus, and (iii) bounded commits. We ablate the two regularizers, mean anchoring ($\beta$) and the entropy bonus ($\tau$), in Section~\ref{sec:ablation} to quantify their effect on the learned prune ratio $p^\star$ and downstream quality.
Mean anchoring in Eq.~\eqref{eq:policy_objective} and Eq.~\eqref{eq:surrogate} penalizes deviations of the policy mean \(\mu\) from the currently committed ratio \(p_{\text{curr}}\), which plays a trust region like role and discourages abrupt shifts in sparsity.
To avoid premature collapse, we maintain exploration by flooring \(\sigma\) for numerical stability and applying the entropy bonus \(\tau\) in the \(\sigma\) update each controller round (Eq.~\eqref{eq:surrogate}).
Finally, the bounded commit in Eqs.~\eqref{eq:delta_p} to \eqref{eq:commit} caps the per-round change in \(p_{\text{curr}}\), reducing mask thrashing and limiting pruning shocks.
Together, importance-based masking that decides what to drop, the controller policy that decides how much to drop, and bounded commits that decide how to change, make \(p\) a learned control variable without changing LoRA placement, rank, or training schedules.

\subsection{Ablation Study}
\label{sec:ablation}

\noindent\textbf{Effect of controller regularization.} We ablate the two stabilization terms introduced in Section~\ref{sec:stab}, namely mean anchoring ($\beta$) and the entropy bonus ($\tau$).
On Arabic XL-Sum, we ablate the mean anchoring weight \(\beta\) and entropy bonus \(\tau\) in the GRPO objective to study how regularization shapes the selected prune ratio and downstream quality.
Table~\ref{tab:abl-kl-ent} reports the selected prune ratio \(p^\star\) together with Arabic BERTScore-F1 and ROUGE-L for each configuration.
The default setting \(\beta{=}0.05\) and \(\tau{=}0.01\) selects \(p^\star{=}64.69\%\) and serves as our reference point for comparison Table~\ref{tab:abl-kl-ent}.
Removing the entropy bonus \(\tau{=}0\) shifts \(p^\star\) toward heavier pruning \(\approx 79\%\) and reduces both F1 and ROUGE-L (Table~\ref{tab:abl-kl-ent}).
Removing both regularizers \(\beta{=}0,\tau{=}0\) yields a similarly high pruning level and also degrades quality relative to the default Table~\ref{tab:abl-kl-ent}.
Increasing the entropy bonus to \(\tau{=}0.05\) selects a lower prune ratio \(p^\star{=}51.14\%\) with F1 close to the default but a small ROUGE-L drop Table~\ref{tab:abl-kl-ent}.
Lowering the anchoring weight to \(\beta{=}0.01\), or dropping anchoring while keeping \(\tau{=}0.01\), leaves \(p^\star\) near \(64\%\) with only minor metric changes Table~\ref{tab:abl-kl-ent}.

\begin{figure}[htb]
  \centering
  \includegraphics[width=\columnwidth]{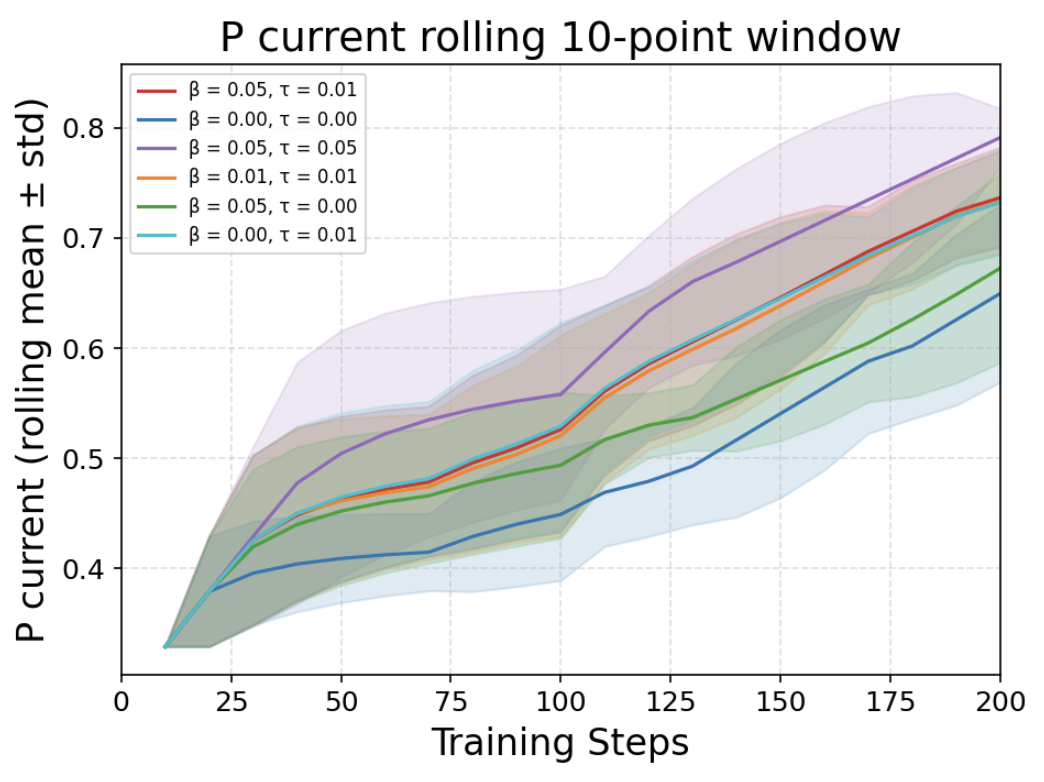}
  \caption{Rolling mean $\pm$ one standard deviation of the committed prune ratio $p_{\text{curr}}$ on Arabic XL-Sum, using a 10-sample window over logged controller records. Narrower shaded bands indicate more stable $p_{\text{curr}}$ trajectories, while wider bands indicate higher variability.}
  \label{fig:p-current-rolling}
\end{figure}

Figure~\ref{fig:p-current-rolling} visualizes how the committed prune ratio evolves during sparsity policy learning.
Each point shows the rolling mean of \(p_{\text{curr}}\) over a 10-sample window of the logged controller records, and the shaded region shows the corresponding \(\pm\) one standard deviation band.
Across settings, wider bands indicate less stable \(p_{\text{curr}}\) trajectories with higher variability across the rolling window.

\newcolumntype{L}{>{\raggedright\arraybackslash}p{0.30\columnwidth}}

\begin{table}[h!]
\centering
\footnotesize                              
\setlength{\tabcolsep}{3pt}                
\renewcommand{\arraystretch}{0.92}         
\begin{tabularx}{0.92\columnwidth}{Lccccc} 
\toprule
\textbf{Variant} & $\beta$ & $\tau$ & $p^\star$ (\%) & \textbf{F1} & \textbf{R-L} \\
\midrule

Default $\beta$ \& $\tau$ & 0.05 & 0.01 & 64.69 & 75.32 & 20.34 \\
w/o $\tau$      & 0.05 & 0.00 & 78.88 & 73.81 & 17.82 \\
High $\tau$    & 0.05 & 0.05 & 51.14 & 75.10 & 19.96 \\
Low $\beta$          & 0.01 & 0.01 & 64.47 & 75.14 & 20.50 \\
w/o $\beta$           & 0.00 & 0.01 & 64.45 & 75.05 & 20.26 \\
w/o $\beta$ \& $\tau$         & 0.00 & 0.00 & 78.54 & 74.01 & 18.34 \\

\bottomrule
\end{tabularx}
\caption{Ablation on GRPO regularization Arabic XL-Sum. $\beta$ = mean anchoring weight, $\tau$ = entropy bonus, $p^\star$ = learned prune ratio. F1 = BERTScore-F1, R-L = ROUGE-L.}
\label{tab:abl-kl-ent}
\end{table}

\noindent\textbf{Micro dev size sensitivity. }
We also probe sensitivity to the micro dev size for the default setting with \(\beta{=}0.05\) and \(\tau{=}0.01\).
Varying the number of micro dev examples from \(m \in \{4, 8, 16, 32\}\) keeps the selected prune ratio within \(64\%\) to \(66\%\) and changes Arabic BERTScore-F1 and ROUGE-L by at most \(0.3\) and \(1.2\) points respectively Table~\ref{tab:microdev}, which suggests that GRASP LoRA remains stable even with very small micro dev slices.

\begin{table}[h]
  \centering
  \footnotesize
  \begin{tabular}{cccc}
    \toprule
    $m$ & $p^\star$ (\%) & F1 & R-L \\
    \midrule
     4  & 64.34 & 75.09 & 20.35 \\
     8  & 66.45 & 75.02 & 19.38 \\
    16  & 64.96 & 75.32 & 20.34 \\
    32  & 64.46 & 75.18 & 20.55 \\
    \bottomrule
  \end{tabular}
  \caption{Effect of micro dev size $m$ on the learned prune ratio and Arabic XL-Sum test performance for the default setting.}
  \label{tab:microdev}
\end{table}

\subsection{Time Efficiency}
Grid search trains a full model once per candidate prune ratio, so cost grows linearly with the grid size. Our method replaces those many full runs with one controller pass on a small micro dev slice and a single final merge and prune run at the learned setting. Table~\ref{tab:time} shows that this yields a \(3.90\times\) to \(7.45\times\) reduction in end to end runtime depending on the dataset and language pair, on identical hardware.

\begin{table}[H]
\centering
\footnotesize
\resizebox{\linewidth}{!}{%
\begin{tabular}{lcccc}
\toprule
\textbf{Method} & \textbf{Dataset} & \textbf{\# runs} & \textbf{runtime (min)} & \textbf{Speedup} \\
\midrule

\multicolumn{5}{l}{\textbf{En + Ar}} \\
Adaptive LoRA  & XL-Sum & 8 & 242 & 1.00$\times$ \\
\textbf{GRASP LoRA } & XL-Sum & \textbf{2} & \textbf{62} & \textbf{3.90$\times$} \\
Adaptive LoRA  & MLQA & 8 & 160 & 1.00$\times$ \\
\textbf{GRASP LoRA } & MLQA & \textbf{2} & \textbf{25} & \textbf{6.40$\times$} \\

\midrule
\multicolumn{5}{l}{\textbf{En + Ch}} \\
Adaptive LoRA  & XL-Sum & 8 & 272 & 1.00$\times$ \\
\textbf{GRASP LoRA } & XL-Sum & \textbf{2} & \textbf{48} & \textbf{5.66$\times$} \\
Adaptive LoRA  & MLQA & 8 & 164 & 1.00$\times$ \\
\textbf{GRASP LoRA } & MLQA & \textbf{2} & \textbf{22} & \textbf{7.45$\times$} \\

\bottomrule
\end{tabular}}
\caption{Runtime comparison on XL-Sum and MLQA: Adaptive LoRA (grid search) vs.\ GRASP LoRA (policy learning + final run).}

\label{tab:time}
\end{table}

\section{Conclusion}
In this paper, we introduced GRASP LoRA, a method that learns how much to prune when merging LoRA adapters for cross lingual transfer. Unlike grid search, GRASP LoRA uses training feedback to select an effective sparsity level, improving both efficiency and accuracy. We validated the approach on two datasets across two target languages with a single backbone model. Future work includes layer wise sparsity and broader evaluation.

\section*{Limitations}
Our evaluation is limited in scope in terms of experimental coverage. 
While we conduct experiments using a single backbone model (Llama 3 8B) 
and one hardware setup, the proposed controller design itself is not 
tied to any specific model architecture, parameterization, or training 
recipe. In principle, the GRASP controller operates on merged adapter 
parameters and validation feedback, and can therefore be applied to 
other backbone families, model scales, and PEFT configurations. 
Extending empirical validation to additional architectures, model sizes, 
and training setups is left for future work.
We also focus on two tasks XL-Sum summarization and MLQA extractive QA and two target languages Arabic and Chinese, with English as the only source language, leaving broader coverage across tasks, domains, and language families for future work. In addition, we evaluate Arabic only in its standard form and do not test dialectal Arabic, so performance under dialectal distribution shifts is unknown. Finally, while we report runtime speedups, we do not measure deployment metrics such as latency, memory footprint, or energy.

\section*{Ethical Considerations}
We train and evaluate GRASP LoRA on XL-Sum and MLQA, which are derived from public news articles and multilingual QA resources. These datasets can contain societal and cultural biases, and in our transfer setting English supervision may carry English-centric framing into Arabic and Chinese outputs. Because our method learns sparsity from task feedback rather than addressing representation, it does not prevent biased content selection or omission in summaries, or biased answer spans in extractive QA. We also do not perform a dedicated bias, toxicity, or fairness evaluation across demographic groups. And we recommend caution and human oversight when using the resulting models in high stakes or sensitive applications.

\bibliography{custom}



\appendix

\section{Additional Results}
\label{app:additional-results}

\subsection{Additional automatic metrics}
Table~\ref{tab:appendix-extra-metrics} reports additional metrics beyond the core ones in Table~\ref{tab:joint-xlsum-mlqa}. For XL-Sum we include ROUGE-1, ROUGE-2, chrF, and BERTScore precision and recall, and for MLQA we compute BLEU, ROUGE, and chrF on the extracted answer strings. These metrics are consistent with the main results and provide a broader view of generation quality.

\subsection{Additional qualitative example}
\begin{figure*}[h!]
  \centering
  \includegraphics[width=\textwidth]{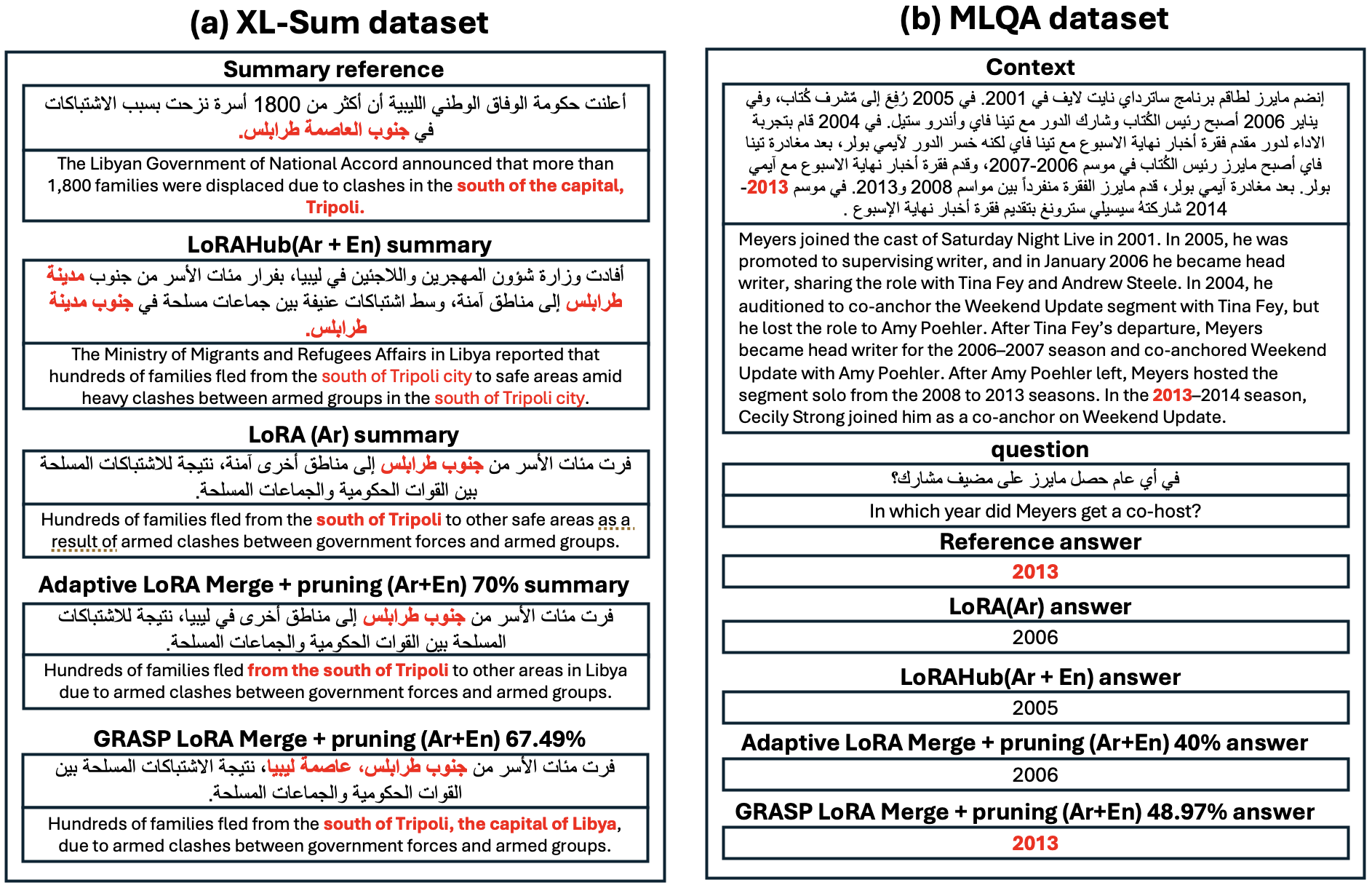}
\caption{Qualitative examples on Arabic transfer. (a) XL-Sum summarization: GRASP LoRA at the learned prune ratio ($p^\star{=}67.49\%$) preserves the key factual detail that Tripoli is the capital of Libya, while LoRA, LoRAHub, and the best grid searched prune ratio omit it. (b) MLQA extractive QA: GRASP LoRA at $p^\star{=}48.97\%$ predicts the correct answer year (2013), while the baselines shown here predict incorrect years. All outputs are generated in Arabic; we show English glosses (approximate translations) under the Arabic text for readability.}

  \label{fig:qual-panel}
\end{figure*}

We provide one additional qualitative comparison for Arabic transfer to complement the aggregate metrics in Table~\ref{tab:joint-xlsum-mlqa}. Figure~\ref{fig:qual-panel} illustrates typical failure modes of the baselines and shows how GRASP LoRA better preserves key factual details and extracts correct answer spans. 

\newcommand{\blockheadExtra}[1]{%
  \addlinespace[0.4em]
  \multicolumn{13}{l}{\textbf{#1} {\scriptsize}}\\[-0.35em]
  \cmidrule(lr){1-13}
}

\begin{table*}[t]
\centering
\normalsize
\setlength{\tabcolsep}{4.5pt}
\renewcommand{\arraystretch}{1.05}
\resizebox{\textwidth}{!}{%
\begin{tabular}{lccccccccccccc}
\hline
 & \multicolumn{6}{c}{\textbf{XL-Sum (extra metrics)}} & \multicolumn{6}{c}{\textbf{MLQA (extra metrics)}} \\
\cmidrule(lr){2-7}\cmidrule(lr){8-13}
\textbf{Type}
& \textbf{Prun.} & \textbf{BERT P} & \textbf{BERT R} & \textbf{ROUGE-1} & \textbf{ROUGE-2} & \textbf{chrF}
& \textbf{Prun.} & \textbf{BLEU-4} & \textbf{ROUGE-1} & \textbf{ROUGE-2} & \textbf{ROUGE-L} & \textbf{chrF} \\
\hline

\blockheadExtra{Without Pruning Arabic}

Zero shot (Ar)
& \textemdash{} & 66.62 & 71.82 & 14.19 & 4.03 & 26.15
& \textemdash{} & 3.49 & 12.22 & 7.79 & 12.13 & 25.96 \\

LoRA (Ar)
& \textemdash{} & 75.17 & 73.60 & 21.78 & 7.13 & 26.37
& \textemdash{} & 32.25 & 50.63 & 25.27 & 50.41 & 43.67 \\

TIES merge (Ar+En)
& \textemdash{} & 67.22 & 72.22 & 15.74 & 4.63 & 27.59
& \textemdash{} & 4.15 & 15.16 & 9.60 & 15.04 & 27.60 \\

LoRAHub (Ar+En)
& \textemdash{} & 74.52 & 73.13 & 20.61 & 6.68 & 25.60
& \textemdash{} & 23.84 & 51.32 & 26.10 & 50.94 & 45.45 \\

\adl{Merge} (Ar+En)
& \textemdash{} & 75.17 & 73.60 & 21.68 & 7.19 & 26.36
& \textemdash{} & 35.73 & 52.63 & 26.72 & 52.53 & 43.51 \\

\blockheadExtra{Best grid search over [10\% -- 80\%] Arabic}

\adl{Merge + pruning} (Ar+En)
& 70
& $75.87 \pm 0.29$
& $74.17 \pm 0.20$
& $24.10 \pm 0.48$
& $9.05 \pm 0.36$
& $27.93 \pm 0.33$
& 40
& $34.78 \pm 0.29$
& $54.98 \pm 0.84$
& $29.96 \pm 0.69$
& $55.01 \pm 1.03$
& $46.89 \pm 0.56$ \\

\blockheadExtra{Learned prune ratio Arabic}

\textbf{Ours \grasp{Merge + pruning} (Ar+En)}
& \textbf{67.49}
& $\textbf{76.58} \pm 0.14$
& $\textbf{75.20} \pm 0.12$
& $\textbf{26.37} \pm 0.30$
& $\textbf{10.74} \pm 0.14$
& $\textbf{30.30} \pm 0.19$
& \textbf{48.97}
& $\textbf{36.60} \pm 0.69$
& $\textbf{57.28} \pm 0.71$
& $\textbf{29.89} \pm 1.39$
& $\textbf{57.49} \pm 0.89$
& $\textbf{47.64} \pm 0.61$ \\

\blockheadExtra{Without Pruning Chinese}

Zero shot (Ch)
& \textemdash{} & 19.66 & 34.19 & 31.17 & 15.99 & 10.92
& \textemdash{} & 5.61 & 14.02 & 11.20 & 13.69 & 27.32 \\

LoRA (Ch)
& \textemdash{} & 35.69 & 27.16 & 35.53 & 20.49 & 10.20
& \textemdash{} & 43.62 & 65.31 & 52.50 & 64.88 & 36.60 \\

TIES merge (Ch+En)
& \textemdash{} & 19.02 & 34.47 & 31.03 & 16.06 & 10.95
& \textemdash{} & 6.77 & 17.08 & 14.36 & 16.69 & 32.88 \\

LoRAHub (Ch+En)
& \textemdash{} & 29.94 & 23.54 & 31.20 & 15.68 & 7.72
& \textemdash{} & 51.11 & 66.12 & 54.65 & 65.70 & 55.69 \\

\adl{Merge} (Ch+En)
& \textemdash{} & 37.31 & 27.73 & 35.38 & 20.98 & 10.28
& \textemdash{} & 52.12 & 67.69 & 55.39 & 67.61 & 54.86 \\

\blockheadExtra{Best grid search over [10\% -- 80\%] Chinese}

\adl{Merge + pruning} (Ch+En)
& 50
& $36.35 \pm 0.25$
& $28.48 \pm 0.41$
& $36.11 \pm 0.31$
& $21.40 \pm 0.13$
& $10.46 \pm 0.24$
& 10
& $55.47 \pm 1.03$
& $70.11 \pm 1.38$
& $57.74 \pm 1.04$
& $70.29 \pm 1.62$
& $59.21 \pm 0.03$ \\

\blockheadExtra{Learned prune ratio Chinese}

\textbf{Ours \grasp{Merge + pruning} (Ch+En)}
& \textbf{56.94}
& $\textbf{36.64} \pm 0.10$
& $\textbf{31.12} \pm 0.18$
& $\textbf{37.75} \pm 0.25$
& $\textbf{22.53} \pm 0.18$
& $\textbf{11.47} \pm 0.07$
& \textbf{23.73}
& $\textbf{58.10} \pm 0.95$
& $\textbf{71.73} \pm 0.30$
& $\textbf{59.85} \pm 0.53$
& $\textbf{71.80} \pm 0.40$
& $\textbf{61.06} \pm 0.70$ \\

\hline
\end{tabular}}
\caption{Additional metrics on XL-Sum and MLQA for Arabic and Chinese. For XL-Sum, we report BERTScore precision and recall, ROUGE-1, ROUGE-2, and chrF. For MLQA, we compute BLEU-4, ROUGE-1/2/L, and chrF on the extracted answer strings, alongside the standard EM and token level QA F1 in Table~\ref{tab:joint-xlsum-mlqa}. All scores are on a 0 to 100 scale; entries of the form $a \pm b$ report the mean $a$ and standard deviation $b$ over three seeds.}

\label{tab:appendix-extra-metrics}
\end{table*}

\section{Implementation Details}
\label{app:implementation-details}

\subsection{Metric configuration}
\label{app:metric-details}

We report the primary metrics in Table~\ref{tab:joint-xlsum-mlqa} and additional metrics in Table~\ref{tab:appendix-extra-metrics}. For \textbf{XL-Sum}, we compute ROUGE and chrF on detokenized model outputs and references, using language-appropriate segmentation. Specifically, ROUGE is computed at the \textbf{word level} for Arabic and at the \textbf{character level} for Chinese to avoid sensitivity to Chinese word segmentation. We report ROUGE-1, ROUGE-2, and ROUGE-L, and we use the same ROUGE configuration across all methods for a given language.

We compute \textbf{SacreBLEU} BLEU-4 with \texttt{tokenize=intl} for Arabic and \texttt{tokenize=zh} for Chinese, following standard multilingual tokenization settings. For \textbf{chrF}, we use the default character n-gram configuration provided by the evaluation toolkit.

We report \textbf{BERTScore} for XL-Sum as precision, recall, and F1, and we use the same multilingual encoder and scoring configuration across all compared methods to ensure comparability. For \textbf{MLQA}, we evaluate extractive predictions as strings: EM and token level QA F1 are computed using the standard MLQA normalization and span-matching procedure. In addition, we compute BLEU, ROUGE, and chrF on the extracted answer strings (Table~\ref{tab:appendix-extra-metrics}) to provide complementary, surface-form similarity signals alongside EM and QA F1.

\subsection{Prompt Templates}
\label{app:prompts}
We use the same prompt structure across languages to isolate the effect of sparsity learning. The Arabic and Chinese templates are direct translations of the English versions and preserve the same fields and ordering.
(\texttt{\{article\}}, \texttt{\{context\}}, \texttt{\{question\}}).

\noindent\textbf{Summarization (English template).}
\begin{quote}
\begin{Verbatim}[fontsize=\small,breaklines=true]
Summarize the following news article.
Article:{article}
Summary:
\end{Verbatim}
\end{quote}

\noindent\textbf{Extractive QA (English template).}
\begin{quote}
\begin{Verbatim}[fontsize=\small,breaklines=true]
Answer the question based on the context.
Context:{context}
Question:{question}
Answer:
\end{Verbatim}
\end{quote}

\subsection{Hyperparameters}
\label{app:hyperparams}
Table~\ref{tab:hyperparams} lists the training, LoRA, and controller hyperparameters shared across all runs. we use these settings for every language and task, and we only vary the controller regularization in the ablations.

\begin{table}[H]
  \centering
  \footnotesize
  \setlength{\tabcolsep}{4pt}
  \renewcommand{\arraystretch}{1.05}
  \begin{tabularx}{\columnwidth}{l >{\raggedleft\arraybackslash}X}
    \toprule
    Hyperparameter & Value \\
    \midrule
    \multicolumn{2}{c}{\textbf{Training and LoRA}} \\
    \midrule
    Training epochs & 10 \\
    Per-device batch size & 1 \\
    Gradient accumulation steps & 1 \\
    Global effective batch size & 1 \\
    Learning rate & $1.0 \times 10^{-4}$ \\
    Optimizer & AdamW \\
    LoRA rank $r$ & 8 \\
    LoRA scaling $\alpha$ & 32 \\
    LoRA dropout & 0.05 \\
    LoRA injection sites & Q and V projections only \\
    \midrule
    \multicolumn{2}{c}{\textbf{GRPO controller}} \\
    \midrule
    Prune range $[p_{\min}, p_{\max}]$ & $[0.10, 0.80]$ \\
    Initial prune ratio $p_{\text{init}}$ & $0.40$ \\
    Controller learning rate $\eta_{\pi}$ & $0.05$ \\
    Controller interval $K$ & 10 steps \\
    Candidates per round $C$ & 3 \\
    micro dev batch size $m$ & 16 \\
    Max commit $\Delta_{\max}$ & 0.10 \\
    \bottomrule
  \end{tabularx}
  \caption{Shared training, LoRA, and controller hyperparameters used across all experiments.}
  \label{tab:hyperparams}
\end{table}

\end{document}